\documentclass[letterpaper,11pt]{article}
\pdfoutput=1
\usepackage{url}
\usepackage{tabto}
\NumTabs{8}
\usepackage{multirow}
\usepackage{adjustbox}
\usepackage{lscape}
\usepackage{natbib}

\usepackage{type1cm}
\usepackage{textcomp}
\usepackage{lettrine}
\usepackage{calc}
\usepackage{hyperref}
\usepackage[T5]{fontenc}
\usepackage[margin=1.5in]{geometry}
\usepackage{amsmath}
\usepackage{amssymb}
\usepackage{mathrsfs}
\usepackage{authblk}
\usepackage{graphicx}
\usepackage[labelfont=bf, labelsep=period, font=small]{caption}
\usepackage{fancyhdr}
\hypersetup{
colorlinks, %
citecolor=black, %
filecolor=black, %
linkcolor=black,%
urlcolor=black,%
}

\title{\endgraf\rule{\textwidth}{3pt}\vspace{4pt}
\textbf{A Fully Convolutional Neural Network for Cardiac Segmentation in Short-Axis MRI}
\endgraf\rule{\textwidth}{1pt}
}
\author{\textbf{Phi Vu Tran}}
\affil{Strategic Innovation Group \\ Booz | Allen | Hamilton \\ McLean, VA USA}
\affil{\normalsize{\texttt{\href{mailto:tran_phi@bah.com}{tran\_phi@bah.com}}}}
\date{}

\begin{document}
\maketitle

\begin{quote}
\textbf{Abstract:} Automated cardiac segmentation from magnetic resonance imaging datasets is an essential step in the timely diagnosis and management of cardiac pathologies. We propose to tackle the problem of automated left and right ventricle segmentation through the application of a deep fully convolutional neural network architecture. Our model is efficiently trained end-to-end in a single learning stage from whole-image inputs and ground truths to make inference at every pixel. To our knowledge, this is the first application of a fully convolutional neural network architecture for pixel-wise labeling in cardiac magnetic resonance imaging. Numerical experiments demonstrate that our model is robust to outperform previous fully automated methods across multiple evaluation measures on a range of cardiac datasets. Moreover, our model is fast and can leverage commodity compute resources such as the graphics processing unit to enable state-of-the-art cardiac segmentation at massive scales. The models and code are available at \url{https://github.com/vuptran/cardiac-segmentation}.


\par
\vspace{10pt}
\end{quote}

\section{Introduction}
\lettrine{C}~ardiovascular diseases are the number one cause of death globally, according to the World Health Organization\footnotemark[1] \footnotetext[1]{\url{http://www.who.int/cardiovascular\_diseases/en/}. Accessed February 8 2016.}. Management of cardiac pathologies typically relies on numerous cardiac imaging modalities, which include echocardiogram, computerized tomography, and magnetic resonance imaging (MRI). The current gold standard is to leverage non-invasive cine MRI to quantitatively analyze global and regional cardiac function through the derivation of clinical parameters such as ventricular volume, stroke volume, ejection fraction, and myocardial mass. Calculation of these parameters depends upon accurate manual delineation of endocardial and epicardial contours of the left ventricle (LV) and right ventricle (RV) in short-axis stacks. Manual delineation is a time-consuming and tedious task that is also prone to high intra- and inter-observer variability \citep{Petitjean:2011, Miller:2013, Tavakoli:2013, Suinesiaputra:2014}. Thus, there exists a need for a fast, accurate, reproducible, and fully automated cardiac segmentation method to help facilitate the diagnosis of cardiovascular diseases.

There are a number of open technical challenges in automated LV and RV segmentation \citep{Petitjean:2011, Tavakoli:2013, Queiros:2014}:

\begin{itemize}
\item The overlap of pixel intensity distributions between cardiac objects and surrounding background structures;
\item The shape variability of the endocardial and epicardial contours across slices and phases;
\item Extreme imbalance in the number of pixels belonging to object class versus background;
\item Fuzzy boundary and edge information, especially in basal and apical slices;
\item Variability in cine MRI from different institutions, scanners, and populations;
\item Inherent noise associated with cine MRI.
\end{itemize}

Although research over the past decade has addressed some of the above technical difficulties to achieve incremental progress on automated ventricle segmentation from short-axis cine MRI, the resulting automated segmentation contours still need to be significantly improved in order to be useable in the clinical setting \citep{Petitjean:2011}. Furthermore, the evaluation of previous research has been limited in scope, on small benchmark datasets that may not represent the real-world variability in image quality and cardiac anatomical and functional characteristics across sites, institutions, scanners, and populations. In addition, previously proposed methods require some \emph{a priori} knowledge about the cardiac ventricles in order to increase their accuracy and robustness \citep{Petitjean:2011}. For semi-automated approaches, direct user interaction is a form of \emph{a priori} knowledge. For fully automated methods, \emph{a priori} information includes hand-engineered features about the spatial relationships of the LV and RV objects and their surrounding structures, knowledge of the heart biomechanics, or anatomical assumptions about the statistical shapes of the objects (e.g., circular geometry of the LV, and complex crescent shape of the RV). Such assumptions about the LV and RV objects, through either weak or strong priors, may contribute to the propensity of previous methods to overfit on a particular training dataset.

\begin{figure}
\centering
\includegraphics[width=\textwidth,height=\textheight,keepaspectratio]{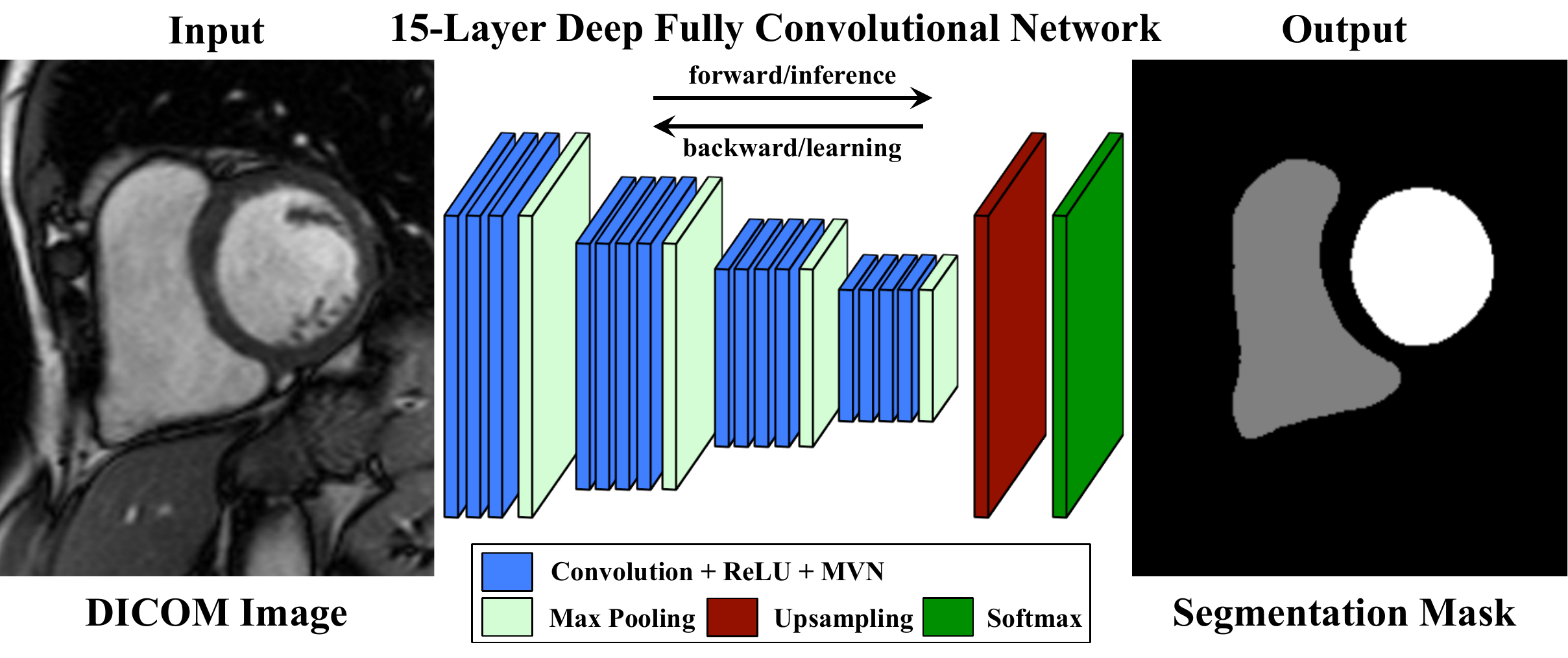}
\centering
\caption{A schematic of our proposed fully convolutional neural network architecture. Acronyms: ReLU -- Rectified Linear Unit; MVN -- Mean-Variance Normalization.}
\label{fig1}
\end{figure}

Our contribution in this paper is a comprehensive evaluation of a convolutional neural network (CNN) architecture on multiple benchmark MRI datasets consisting of the left and right ventricle. The basic components of a CNN architecture include trainable filters that can automatically learn intricate features and concepts from a training set in a supervised manner, without the need for feature engineering or to hard-code \emph{a priori} knowledge. CNNs are also amenable to transfer learning \citep{DeCAF:2013, Zeiler:2014, Oquab:2014, Yosinski:2014, Razavian:2014}, a task that proves to be valuable in the absence of abundant training data. We show that the creative application of a CNN variant, the \emph{fully convolutional neural network} (FCN), achieves state-of-the-art semantic segmentation in short-axis cardiac MRI acquired at multiple sites and from different scanners. The proposed FCN architecture is efficiently trained end-to-end on a graphics processing unit (GPU) in a single learning stage to make inference at every pixel, a task commonly known as pixel-wise labeling or per-pixel classification. At test time, the model independently segments each image in milliseconds, so it can be deployed in parallel on clusters of CPUs, GPUs, or both for scalable and accurate ventricle segmentation. To our knowledge, this is the first application of a CNN architecture for pixel-wise labeling in cardiac MRI.

The remainder of this paper is structured as follows: Section 2 briefly surveys previous research on LV and RV segmentation and fully convolutional neural networks. Section 3 presents an experimental framework to demonstrate and evaluate the efficacy of our FCN model on a range of publicly available benchmark cardiac MRI datasets. Section 4 reports results and analysis of the evaluation. Finally, Section 5 concludes with a summary and parting remarks.

\section{Previous Work}

\subsection{Left Ventricle Segmentation}
The task of delineating the left ventricle endocardium and epicardium from cine MRI throughout the cardiac cycle has received much research focus and attention over the past decade. Two grand challenges, Medical Image Computing and Computer Assisted Intervention (MICCAI) 2009 LV Segmentation Challenge \citep{Radau:2009} and Statistical Atlases and Computational Modeling of the Heart (STACOM) 2011 LV Segmentation Challenge \citep{Suinesiaputra:2014}, have emerged with the goal of advancing the state of the art in automated LV segmentation. To facilitate research and development in this arena, the challenges provide benchmark datasets that come with expert ground truth contours and standard evaluation measures to assess automated segmentation performance. \citet{Petitjean:2011} provide a comprehensive survey of previous methods for LV segmentation that include multi-level Otsu thresholding, deformable models and level sets, graph cuts, knowledge-based approaches such as active and appearance shape models, and atlas-based methods. Although these methods have achieved limited success on small benchmark LV datasets, they suffer from low robustness, low accuracy, and limited capacity to generalize over subjects with heart conditions outside of the training set.

More recently, \citet{Ngo:2013} couple Restricted Boltzmann Machines and a level set method to produce competitive results on a small benchmark LV dataset \citep{Radau:2009}. However, their method is semi-automated that requires user input. \citet{Queiros:2014} propose a novel automated 3D+time LV segmentation framework that combines automated 2D and 3D segmentation with contour propagation to yield accurate endocardial and epicardial contours on the same LV dataset. \citet{Avendi:2015} integrate recent advances in machine learning such as stacked autoencoders and convolutional neural networks with a deformable model to establish a new state of the art on LV endocardium segmentation. The main limitation of the recent methods is that they are multi-stage approaches that require manual offline training and extensive hyper-parameter tuning, which can be cumbersome. Furthermore, the evaluation of the methods by \citet{Ngo:2013} and \citet{Avendi:2015} is limited to only LV endocardial contours. None of these methods is evaluated for the task of automated right ventricle segmentation.

\subsection{Right Ventricle Segmentation}
The task of delineating the right ventricle endocardium and epicardium from short-axis cine MRI at various phases of the cardiac cycle shares similar goals, clinical motivations, and inherent technical difficulties as its LV counterpart. However, it does not receive as much research attention, partly because RV segmentation algorithms have never had access to a common database with expert ground truth contours. In an effort to advance the research and development of RV segmentation towards clinical applications, the MICCAI 2012 Right Ventricle Segmentation Challenge proposes a benchmark MRI dataset that comes with expert ground truth segmentation contours and standard evaluation measures, following the formats of prior LV segmentation competitions. \citet{Petitjean:2015} survey the automated and semi-automated approaches presented by the seven challenger teams that include three atlas-based methods, two prior-based methods, and two prior-free, image-driven methods that make use of the temporal dimension of the data. This competition highlights the current interest in methods based on the multi-atlas segmentation framework \citep{Rohlfing:2004, Klein:2005, Heckemann:2006}, which is becoming one of the most widely used and successful image segmentation techniques in medical imaging applications \citep{Iglesias:2015}.

Although the methods presented in the MICCAI 2012 challenge achieve reasonable segmentation accuracy, there is still much room left for improvement, especially if the methods are to be utilized in the clinical setting. The main limitation of previous methods based on statistical shape modeling, feature engineering, and multi-atlas registration is that they tend to overfit on one particular dataset and may not generalize well to other datasets nor are amenable to transfer learning.

\subsection{Fully Convolutional Neural Networks} 
Convolutional neural networks (CNNs) continue to achieve record-breaking accuracy performances on many visual recognition benchmarks across research domains. CNNs are supervised models trained end-to-end to learn hierarchies of features automatically--without resorting to sometimes complicated input preprocessing, output postprocessing, and feature engineering schemes--yielding robust classification and regression performances. Recent successes of deep CNN architectures like AlexNet \citep{Krizhevsky:2012}, VGGNet \citep{Simonyan:2015}, GoogLeNet \citep{Ioffe:2015}, and ResNet \citep{He:2015} have made CNNs the \emph{de facto} standard for whole-image classification. In addition, high-performance deep CNNs have been adapted to advance the state of the art on other visual recognition tasks such as bounding box object detection \citep{Girshick:2014, Girshick:2015, Ren:2016} and semantic segmentation \citep{Long:2015, Liu:2015, Zheng:2015}.

A standard deep CNN architecture for whole-image classification typically consists of convolution layer, nonlinear activation function, pooling layer, and fully connected layer as basic building blocks. \citet{Long:2015} adapt and extend deep classification architectures by removing the fully connected layers and introducing fractional convolution layers to learn per-pixel labels end-to-end from whole-image inputs and corresponding whole-image ground truths. \citet{Long:2015} describe their fractional convolution layer as useful for learning (nonlinear) upsampling filters in order to map or connect coarse outputs to the dense pixel space. Their key to success is to leverage large-scale image classification \citep{Deng:2009} as supervised pre-training, and fine-tune \emph{fully convolutionally} via transfer learning. Others expand upon the FCN idea by adding global context \citep{Liu:2015} and coupling Conditional Random Field learning \citep{Zheng:2015} to push the performance boundaries in semantic segmentation.

\section{Experimental Framework}
We propose to tackle the problem of automated LV and RV segmentation through the application of an FCN architecture. Numerical experiments demonstrate that our FCN model is robust to outperform previous methods across multiple evaluation measures on a range of cardiac MRI datasets. All MRI datasets are in anonymized DICOM format.

\subsection{Datasets}

\textbf{Sunnybrook Cardiac Data \citep{Radau:2009}:} The Sunnybrook dataset comprises cine MRI from 45 patients, or cases, having a mix of cardiac conditions: healthy, hypertrophy, heart failure with infarction, and heart failure without infarction. Expert manual segmentation contours for the endocardium, epicardium, and papillary muscles are provided for basal through apical slices at both end-diastole (ED) and end-systole (ES) phases. This dataset was made available as part of the MICCAI 2009 challenge on automated LV segmentation from short-axis cardiac MRI. The Sunnybrook dataset is available through the Cardiac Atlas Project (CAP)\footnotemark[2] with a public domain license.

\footnotetext[2]{\url{http://www.cardiacatlas.org/studies/sunnybrook-cardiac-data/}}

The Sunnybrook dataset is divided into three disjoint sets of 15 cases each: training, validation, and online. Ground truth contours are provided for training, validation, and online sets. We use the training set to train an FCN model for LV endocardium and epicardium segmentation, and evaluate model performance on the validation and online sets. We do not investigate the segmentation of the papillary muscles because few researchers have done so in the past and therefore it is hard to compare results. It is important to note that we take great care to perform model selection and hyper-parameter tuning on a \emph{development} subset derived from randomly splitting the training set into 0.9/0.1 folds. This procedure is standard protocol to ensure that we do not peek into the validation and online sets that can result in overfitting, and that we stay consistent with how the challenge was conducted. \\

\noindent \textbf{Left Ventricle Segmentation Challenge \citep{Suinesiaputra:2014}:} This dataset, denoted here as LVSC, was made publicly available as part of the STACOM 2011 challenge on automated LV myocardium segmentation from short-axis cine MRI. The dataset is derived from the DETERMINE cohort \citep{Kadish:2009} consisting of 200 patients with coronary artery disease and myocardial infarction. The LVSC dataset comes with expert-guided semi-automated segmentation contours for the myocardium, a region composed of pixels inside the epicardium and outside the endocardium, derived from the Guide-Point Modeling technique \citep{Li:2010}. This approach involves expert input to refine the segmentation contours by interactively positioning a small number of guide points on a subset of slices and frames. The contours are available for basal through apical slices at both ED and ES phases. The LVSC dataset can be downloaded from the LV Segmentation Challenge website via the CAP\footnotemark[3].

\footnotetext[3]{\url{http://www.cardiacatlas.org/challenges/lv-segmentation-challenge/}}

The LVSC dataset is divided into two disjoint sets of 100 cases each: testing and validation. We use the testing set with the provided expert-guided contours to train an FCN model to segment the LV myocardium, and evaluate model performance on the validation set. We split the testing set into 0.95/0.05 training/development folds for experimentation, model selection, hyper-parameter tuning. There are no absolute ground truth contours for the validation set. Instead, the challenge organizers estimate reference consensus images from a set of five segmentation algorithms (two fully automated and three semi-automated requiring user input) using the STAPLE (Simultaneous Truth and Performance Level Estimation) method \citep{Warfield:2004}. The idea is to establish a large community resource of ground truth images based on common data for the development, validation, and benchmarking of LV segmentation algorithms \citep{Suinesiaputra:2014}. Reference consensus images are not provided for the validation set, so we submit our predicted myocardial contours to the challenge organizers for independent evaluation. \\

\noindent \textbf{Right Ventricle Segmentation Challenge \citep{Petitjean:2015}:} This dataset, denoted here as RVSC, was provided as part of the MICCAI 2012 challenge on automated RV endocardium and epicardium segmentation from short-axis cine MRI. The dataset comprises 48 cases having various cardiac pathologies. Expert manual endocardial and epicardial contours are provided for basal through apical slices at both ED and ES phases. The RVSC dataset is available for download from the LITIS lab at the University of Rouen\footnotemark[4].

\footnotetext[4]{\url{http://www.litislab.fr/?projet=1rvsc}}

The RVSC dataset is divided into three disjoint sets of 16 cases each: training, test1, and test2. Expert manual contours are provided for the training set only. We split the training set into 0.9/0.1 training/development subsets for experimentation, model selection, and hyper-parameter tuning. At test time, we submit our predicted RV endocardial and epicardial contours for test1 and test2 sets to the challenge organizers for independent evaluation. \\


\subsection{Data Preparation and Augmentation}
We observe that the heart cavity containing both the left and right ventricle appears roughly at the center of each short-axis slice. We proceed to take the square center crop of each image to define our region of interest (ROI). The size of the ROI can have an impact on the accuracy performance of the FCN model. We choose a multi-resolution approach to crop the ROI at multiple sizes that wholly contain the ventricles. Multi-scale cropping provides the following benefits:

\begin{itemize}
\item Augment the training set by providing multiple views of the same image at multiple resolutions;
\item Capture the ROI while providing a ``zooming'' effect for enhanced feature learning;
\item Mitigate class imbalance by removing unnecessary background pixels;
\item Accelerate computations via the reduction of input spatial dimensions.
\end{itemize}

The 16-bit MRI datasets have a wide range of pixel intensities that directly influence the accuracy performance of automated segmentation models, especially if the acquired images come from multiple sites using different scanner types or manufacturers. We normalize the pixel intensity distribution of each input image by subtracting its mean and dividing the resulting difference by its standard deviation. The normalized output is an image with pixel values having zero mean and unit variance. Mean-variance normalization (MVN) is a simple yet effective technique that significantly enhances the learning capacity of our FCN model during training and segmentation performance during testing across datasets. No further preprocessing is done on the input image pixels. We also perform affine transformations (rotation, vertical flipping, and horizontal flipping) to augment the training set in an effort to mitigate overfitting and further improve model generalization. Multi-scale center cropping and affine transformations artificially inflate the training set by 12-fold, although the resulting augmented dataset is highly correlated. Table~\ref{tab1} summarizes our data augmentation strategy for each dataset.

\begin{table}[ht]
\captionsetup{font=small}
\begin{center}
\captionof{table}{Data augmentation strategy for each dataset during training. At test time, we standardize input images via center cropping and mean-variance normalization. The tuple $(h, w)$ denotes the height and width of the input image, respectively.}
\begin{adjustbox}{width=1\textwidth}
	\begin{tabular} {| c | c | c | c | c | c |}
	\hline
	& \multicolumn{4}{ c |}{\textbf{Training}} & \textbf{Testing} \\ \hline
	\textbf{Dataset} & \textbf{Center Crop} & \textbf{Rotation} & \textbf{Vertical Flip} & \textbf{Horizontal Flip} & \textbf{Center Crop} \\ \hline
	Sunnybrook & $dim \in [100, 120]$ & $k \times 90, k \in [1, 2, 3]$ & Yes & Yes & $dim = 100$\\ \hline
	LVSC & $dim = \text{int} (\min(h, w) \times 0.6)$ & No & No & No & $dim = \text{int}(\min(h, w) \times 0.6)$ \\ \hline
	RVSC & $dim \in [200, 216]$ & $k \times 90, k \in [1, 2, 3]$ & Yes & Yes & $dim = 200$\\
	\hline
	\end{tabular}
	\label{tab1}
\end{adjustbox}
\end{center}
\end{table}

\subsection{FCN Architecture}
Figure~\ref{fig1} illustrates our proposed FCN architecture, which is selected via cross-validation on a development set. The FCN architecture comprises 15 stacked convolution layers and three layers of overlapping, two-pixel strided max pooling. Each convolution layer is followed by Rectified Linear Unit (ReLU) activation function \citep{Nair:2010} and MVN operation. The architecture has roughly 11 million parameters to be estimated. Such a high-dimensional model is prone to overfit on the relatively small MRI datasets under consideration; we take great care to mitigate overfitting through data augmentation during data preparation, and dropout and regularization during training.

We employ the ``skip'' architecture of \citet{Long:2015} to combine coarse semantic information at deep layers and fine appearance information at shallow layers to learn filters for output upsampling. The end result is a dense heatmap predicting class membership of each pixel in the input image. For time benchmarking purposes, the FCN model takes an average of 61 milliseconds to segment one image of $256 \times 256$ pixels using a single NVIDIA GeForce GTX TITAN X GPU. We apply and evaluate a single instantiation of this FCN architecture on all MRI datasets under consideration.


\subsection{Training Protocol}
We leverage the Caffe deep learning framework \citep{Jia:2014} for the design, implementation, and experimentation of our deep FCN architecture. We employ stochastic gradient descent with momentum of 0.9 to minimize the multinomial logistic loss on per-pixel softmax probabilities from whole-image inputs and ground truths. We randomly initialize parameter weights according to the ``Xavier'' scheme \citep{Glorot:2010}. We further combat the adverse effects of overfitting by using dropout ratio of 0.5 \citep{Srivastava:2014} and $L_2$ weight decay regularization of 0.0005. We train for 10 epochs, or passes over the training set, and anneal the learning rate according to the polynomial decay of the form: $\text{base\_lr} \times \left(1 - \frac{\text{iter}} {\text{max\_iter}}\right) ^ \text{power}$, where $\text{base\_lr} = 0.01$ is the initial learning rate, iter is the current iteration, max\_iter is the dataset-specific maximum number of iterations approximately equal to 10 epochs, and $\text{power} = 0.5$ controls the rate of decay.

\subsection{Transfer Learning}
We also explore the benefits of transfer learning in training deep FCN models with limited data. We first train an FCN model using ``Xavier'' random initialization on the relatively large LVSC testing set of roughly 22,000 DICOM images that come with expert-guided semi-automated contours to obtain a set of learned convolution filters. We call the pre-trained FCN model the source model, the LVSC testing set the source dataset, and the task of estimating LV contours the source task. In transfer learning, we initialize a second FCN model (the target model) with the learned weights from the source model by copying or transferring from selected convolution and upsampling layers. The remaining layers of the target model are then randomly initialized and trained toward a target task using a target dataset via supervised fine-tuning.

We experiment with transferring the learned feature representation from the source task of LV segmentation on the LVSC dataset to the target task of LV segmentation on the Sunnybrook dataset and to the target task of RV segmentation on the RVSC dataset. The transferred weights serve as supervised pre-training that enable training a large target model on small target datasets without severe overfitting. In transfer learning, we set the initial learning rate to be small, $\text{base\_lr} = 0.001$, in order to refine the update of the learned weights during backpropagation. Transfer learning and supervised fine-tuning offer the following benefits:
\begin{itemize}
\item Domain adaptation -- transfer learning allows the learned model on the source task to be adapted to a different, but related, target task. For example, we can enable a source model that learns LV contours to train on estimating RV contours. \citet{Yosinski:2014} document that the transferability of features decreases as the distance between the source task and target task increases, but that transferring features even from distant tasks can be better than using random features;
\item The source and target datasets need not be from the same distribution. For example, the Sunnybrook Cardiac Dataset follows a different distribution than the LVSC dataset because they were acquired from different institutions, even though both datasets represent the LV object;
\item Better convergence and accuracy performance even with limited training data.
\end{itemize}

\section{Empirical Evaluation}

\subsection{Metrics}
Let $a$ and $m$ be the predicted (automated) and ground truth (manual) contours delineating the object class in short-axis MRI, respectively. Let $A$ and $M$ be the corresponding areas enclosed by contours $a$ and $m$, respectively. The following evaluation metrics are used to assess the accuracy of automated segmentation methods using the ground truth as reference. Different challenges use different measures for the respective dataset; we provide an overview of the main metrics reported in the literature for comparative purposes. \\

\noindent \textbf{Sunnybrook Cardiac Dataset}:
\begin{itemize}
\item Average perpendicular distance (APD) measures the distance between contours $a$ and $m$, averaged over all contour points. A high value implies that the two contours \emph{do not} closely match \citep{Radau:2009}. APD is computed in millimeter with spatial resolution obtained from the \texttt{PixelSpacing} DICOM field.
\item The Dice index \citep{Dice:1945} is a measure of overlap or similarity between two contour areas, and is defined as:
\begin{equation*}
\mathcal{D}(A, M) = 2 \frac{A \cap M}{A + M}.
\end{equation*}
The Dice index varies from zero (total mismatch) to unity (perfect match).
\item Percentage of good contours is a fraction of the predicted contours, out of the total number of contours, that have APD less than 5 millimeters away from the ground truth contours \citep{Radau:2009}.
\end{itemize}

\noindent \textbf{LVSC Dataset}:
\begin{itemize}
\item Sensitivity ($p$), specificity ($q$), positive predictive value ($PPV$), and negative predictive value ($NPV$) are defined as:
\begin{equation*}
p = \frac{T_1}{N_1}, \quad q = \frac{T_0}{N_0}, \quad PPV = \frac{T_1}{T_1 + F_1}, \quad NPV = \frac{T_0}{T_0 + F_0},
\end{equation*}
where $T_1$ and $T_0$ are the number of correctly predicted pixels as belonging to the object and background class, while $F_1$ and $F_0$ are the number of misclassified pixels as object and background, respectively. The total number of object and background pixels are denoted by $N_1$ and $N_0$, respectively \citep{Suinesiaputra:2014}.
\item The Jaccard index \citep{Jaccard:1912} is a measure of overlap or similarity between two contour areas, and is defined as:
\begin{equation*}
\mathcal{J}(A, M) = \frac{A \cap M}{A \cup M} = \frac{A \cap M}{A + M - \left(A \cap M \right)}
\end{equation*}
Similar to the Dice index, the Jaccard index varies from zero to unity, with unity representing perfect correspondence with the ground truth.
\end{itemize}

\noindent \textbf{RVSC Dataset}:
\begin{itemize}
\item The Hausdorff distance is a symmetric measure of distance between two contours \citep{Huttenlocher:1993}, and is defined as:
\begin{equation*}
\mathcal{H}(a, m) = \max \left( \max_{i \in a} \left( \min_{j \in m} d(i, j) \right), \max_{j \in m} \left( \min_{i \in a} d(i, j) \right) \right),
\end{equation*}
where $d(\cdot, \cdot)$ denotes Euclidean distance. Similar to APD, a high Hausdorff value implies that the two contours \emph{do not} closely match. The Hausdorff distance is computed in millimeter with spatial resolution obtained from the \texttt{PixelSpacing} DICOM field.
\item The Dice index $\mathcal{D}(A, M)$ as defined above.
\end{itemize}

%

\subsection{Results and Analysis}
\begin{figure}[ht]
\centering
\includegraphics[width=\textwidth,height=\textheight,keepaspectratio]{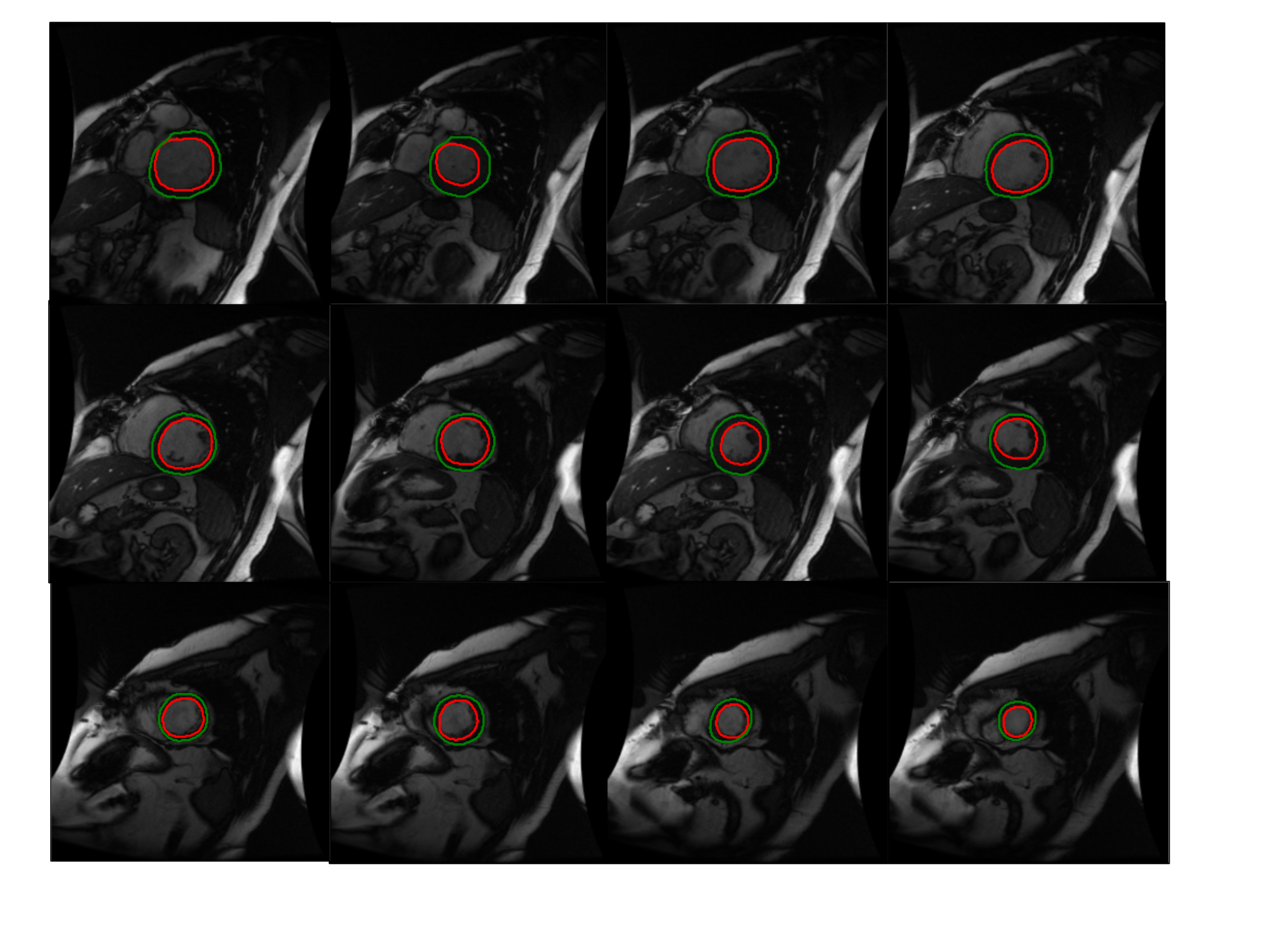}
\centering
\caption{FCN segmentation result of an example test case in the Sunnybrook dataset for both ED and ES phases. Colors: red -- endocardium; green -- epicardium.}
\label{fig2}
\end{figure}

Tables~\ref{tab2},~\ref{tab3}, and~\ref{tab4} summarize our automated segmentation results and compare them to the previous state of the art. On the combined Sunnybrook validation and online sets, our FCN model achieves comparable Dice index with that of the method by \citet{Avendi:2015} for automated LV endocardium segmentation. For all other evaluation measures, our model obtains the best scores across the board. Note that fine-tuning the FCN model does have an accuracy improvement over the same FCN model initialized with random features, a result that has been consistently corroborated in many previous studies. At test time, our model segments the endocardium and epicardium in both validation and online sets (a total of 830 images) in less than 25 seconds. Figure~\ref{fig2} illustrates example predicted endocardial and epicardial contours for the left ventricle using the Sunnybrook dataset.

\begin{table}[ht]
\captionsetup{font=small}
\begin{center}
\captionof{table}{Comparison of LV endocardium and epicardium segmentation performance between our proposed FCN model and previous research using the Sunnybrook Cardiac Dataset. We distinguish performance of the FCN model through either transfer learning and supervised fine-tuning from the source LVSC dataset or Xavier random initialization. Number format: mean value (standard deviation).}
\begin{adjustbox}{width=1\textwidth}
	\begin{tabular} { l  c  c  c  c  c  c  c }
	\hline \\
	\multicolumn{1}{l}{\multirow{2}{*}{\textbf{Method}} } &
	\multicolumn{1}{c}{\multirow{2}{*}{\textbf{\#}\footnotemark[5]}} &
	\multicolumn{2}{c}{\multirow{1}{*}{\textbf{Dice Index}}} &
	\multicolumn{2}{c}{\multirow{1}{*}{\textbf{APD\footnotemark[6] (mm)}}} &
	\multicolumn{2}{c}{\multirow{1}{*}{\textbf{Good Contours (\%)}}} \\
	& & \textbf{Endo} & \textbf{Epi} & \textbf{Endo} & \textbf{Epi} & \textbf{Endo} & \textbf{Epi} \\ \\ \hline \hline
	Our FCN model w/ finetune  & 30
							& 0.92 (0.03)
							& \textbf{0.96 (0.01)}
							& \textbf{1.73 (0.35)}
							& \textbf{1.65 (0.31)}
							& \textbf{98.48 (4.06)}
							& \textbf{99.17 (2.20)} \\
	Our FCN model w/ Xavier init & 30
				 		& 0.92 (0.03)
						& 0.95 (0.02)
						& 1.74 (0.43)
						& 1.69 (0.34)
						& 97.42 (5.86)
						& 98.00 (3.78) \\
	\citep{Avendi:2015} & 30
				        & \textbf{0.94 (0.02)}
				        & --
				        & 1.81 (0.44)
				        & --
				        & 96.69 (5.7)
				        & -- \\
	\citep{Queiros:2014} & 45
					 & 0.90 (0.05)
					 & 0.94 (0.02)
					 & 1.76 (0.45)
					 & 1.80 (0.41)
					 & 92.70 (9.5)
					 & 95.40 (9.6) \\
	\citep{Ngo:2013} & 45
					        & 0.90 (0.03)
					        & --
					        & 2.08 (0.40)
					        & --
					        & 97.91 (6.2)
					        & -- \\
	\citep{Hu:2013} & 45
				 & 0.89 (0.03)
				 & 0.94 (0.02)
				 & 2.24 (0.40)
				 & 2.19 (0.49)
				 & 91.06 (9.4)
				 & 91.21 (8.5) \\
	\citep{Liu:2012} & 45
				  & 0.88 (0.03)
				  & 0.94 (0.02)
				  & 2.36 (0.39)
				  & 2.19 (0.49)
				  & 91.17 (8.5)
				  & 90.78 (10.7) \\
	\citep{Huang:2011} & 45
				       & 0.89 (0.04)
				       & 0.93 (0.02)
				       & 2.16 (0.46)
				       & 2.22 (0.43)
				       & 79.20 (19.0)
				       & 83.90 (16.8) \\
	\citep{Constantinides:2009} & 30
						    & 0.89 (0.04)
						    & 0.92 (0.02)
						    & 2.04 (0.47)
						    & 2.35 (0.57)
						    & 90.35
						    & 92.56 \\
	\citep{Jolly:2009} 		& 30
				       & 0.88 (0.04)
				       & 0.93 (0.02)
				       & 2.26 (0.59)
				       & 1.97 (0.48)
				       & 95.62 (8.8)
				       & 97.29 (5.8) \\
	
	\hline
	\end{tabular}
	\label{tab2}
\end{adjustbox}
\end{center}
\vspace{-8pt}
{\small
\footnotemark[5] Number of test cases: 30 -- validation and online cases; 45 -- training, validation, and online cases. \\
\footnotemark[6] Average Perpendicular Distance.}
\end{table}

\begin{table}[ht]
\captionsetup{font=small}
\begin{center}
\captionof{table}{Comparison of LV myocardium segmentation performance between our proposed FCN model and previous research using the LVSC validation set based on the \texttt{CS*} consensus. Values are taken from Table 2 of \citet{Suinesiaputra:2014}. Number format: mean value (standard deviation).}
\begin{adjustbox}{width=\textwidth}
	\begin{tabular} { l c c c c c c }
	\hline
	\textbf{Method} &
	\textbf{FA/SA}\footnotemark[7] &
	\textbf{Jaccard Index} &
	\textbf{Sensitivity} &
	\textbf{Specificity} &
	\textbf{PPV} &
	\textbf{NPV} \\ \hline \hline
	Our FCN model & FA
				& \textbf{0.74 (0.13)}
				& 0.83 (0.12)
				& \textbf{0.96 (0.03)}
				& 0.86 (0.10)
				& \textbf{0.95 (0.03)} \\
	\citep{Jolly:2012} & FA
				    & 0.69 (0.23)
				    & 0.74 (0.23)
				    & 0.96 (0.05)
				    & \textbf{0.87 (0.16)}
				    & 0.89 (0.09) \\ 
	\citep{Margeta:2012} & FA
					 & 0.43 (0.10)
					 & \textbf{0.89 (0.17)}
					 & 0.56 (0.15)
					 & 0.50 (0.10)
					 & 0.93 (0.09) \\ \hline
	 \citep{Li:2010} (Expert guided) & SA
				& \textbf{0.84 (0.17)}
				& \textbf{0.89 (0.13)}
				& \textbf{0.96 (0.06)}
				& \textbf{0.91 (0.13)}
				& \textbf{0.95 (0.06)} \\
	\citep{Fahmy:2012} & SA
				       & 0.74 (0.16)
				       & 0.88 (0.15)
				       & 0.91 (0.06)
				       & 0.82 (0.12)
				       & 0.94 (0.06) \\
	\citep{Ourselin:2000} & SA
					& 0.64 (0.18)
					& 0.80 (0.17)
					& 0.86 (0.08)
					& 0.74 (0.15)
					& 0.90 (0.08) \\
	\hline
	\end{tabular}
	\label{tab3}
\end{adjustbox}
\end{center}
\vspace{-8pt}
{\small
\footnotemark[7] Fully Automated / Semi-Automated
}
\end{table}

\begin{table}[ht]
\captionsetup{font=small}
\begin{center}
\captionof{table}{Comparison of RV endocardium and epicardium segmentation performance between our proposed FCN model and previous research using the RVSC dataset. We distinguish performance of the FCN model through either transfer learning and supervised fine-tuning from the source LVSC dataset or Xavier random initialization. Values are averaged over test1 and test2 sets in format: mean value (standard deviation).}
\begin{adjustbox}{width=\textwidth}
	\begin{tabular} { l  c  c  c  c  c  c }
	\hline \\
	\multicolumn{1}{l}{\multirow{2}{*}{\textbf{Method}} } &
	\multicolumn{1}{c}{\multirow{2}{*}{\textbf{FA/SA}\footnotemark[7]}} &
	\multicolumn{2}{c}{\multirow{1}{*}{\textbf{Dice Index}}} &
	\multicolumn{2}{c}{\multirow{1}{*}{\textbf{Hausdorff Dist (mm)}}} \\
	& & \textbf{Endo} & \textbf{Epi} & \textbf{Endo} & \textbf{Epi} \\ \\ \hline \hline
	Our FCN model w/ finetune & FA & \textbf{0.84 (0.21)} & \textbf{0.86 (0.20)} & \textbf{8.86 (11.27)} & \textbf{9.33 (10.79)} \\
	Our FCN model w/ Xavier init & FA & 0.80 (0.27) & 0.84 (0.24) & 11.41 (15.25) & 11.27 (15.04) \\
	\citep{Zuluaga:2013} & FA & 0.76 (0.25) & 0.80 (0.22) & 11.51 (10.06) & 11.82 (9.38) \\
	\citep{Wang:2012} & FA & 0.59 (0.34) & 0.63 (0.35) & 25.32 (22.66) & 24.43 (22.26) \\
	\citep{Ou:2012} & FA & 0.58 (0.31) & 0.63 (0.27) & 19.12 (14.39) & 18.85 (13.47) \\ \hline
	\citep{Grosgeorge:2013} & SA & \textbf{0.79 (0.18)} & \textbf{0.84 (0.12)} & \textbf{8.63 (4.54)} & \textbf{9.36 (4.58)} \\
	\citep{Bai:2013} & SA & 0.77 (0.22) & 0.82 (0.16) & 9.52 (5.26) & 9.99 (5.18) \\
	\citep{Maier:2012} & SA & 0.79 (0.22) & -- & 10.47 (6.00) & -- \\
	\citep{Nambakhsh:2013} & SA & 0.58 (0.24) & -- & 21.21 (9.71) & -- \\
	\hline
	\end{tabular}
	\label{tab4}
\end{adjustbox}
\end{center}
\vspace{-8pt}
{\small
\footnotemark[7] Fully Automated / Semi-Automated
}
\end{table}

For the task of predicting myocardial contours on the LVSC validation set, our FCN model achieves the best scores in three out of five metrics including the Jaccard index, specificity, and negative predictive value in comparison to previous fully automated methods. Note that we compare our segmentation results against previous methods based on Table 2 of \citet{Suinesiaputra:2014} using the \texttt{CS*} consensus.

When compared against the expert-guided semi-automated method of \citet{Li:2010}, which was used to generate reference ground truth contours for the LVSC testing set, the FCN model performs significantly worse. It is important to note that the FCN model segments each DICOM image independently using the contextual cues of the image pixels as features, while the Guide-Point Modeling technique of \citet{Li:2010} requires human expert input to refine and approve the segmentation results for all slices and for all frames. There are several difficult cases where our model cannot detect the presence of the LV object in apical/basal slices, mainly because they exhibit ambiguous or imperceptible object boundaries. These cases necessitate user intervention to improve segmentation, which is the reason why the Guide-Point Modeling technique performs so well. However, the main limitation of the guide-point approach is the slow processing time associated with user intervention, giving rise to the crux of the problem in scalability. In contrast, the FCN model is fully automated and scales to massive datasets; at test time, our model segments all 29,859 short-axis images, for 100 cases total, in the LVSC validation set in under 19 minutes.

\begin{figure}[ht]
\centering
\includegraphics[width=\textwidth,height=\textheight,keepaspectratio]{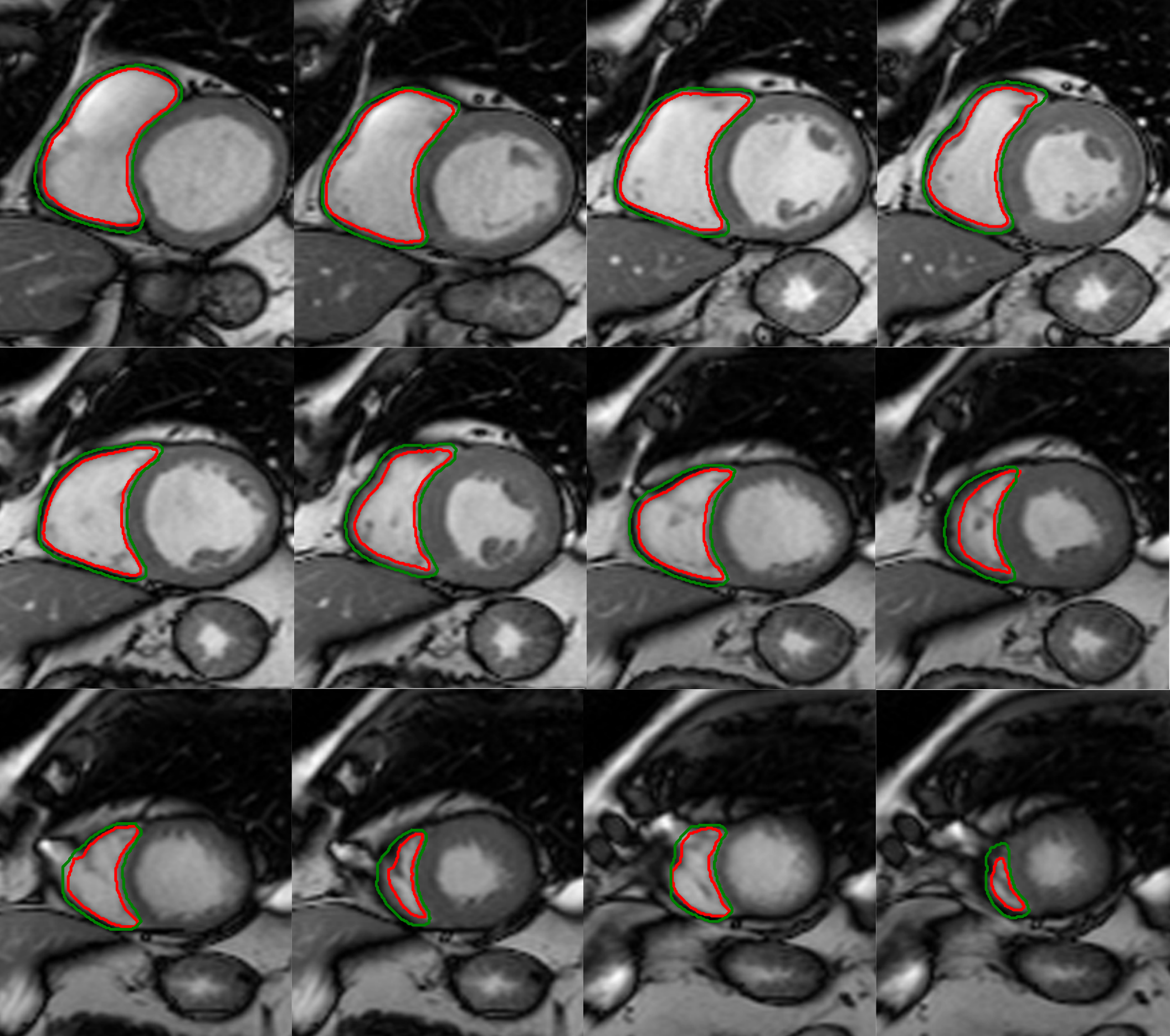}
\centering
\caption{FCN segmentation result of an example test case in the RVSC dataset for both ED and ES phases. Colors: red -- endocardium; green -- epicardium.}
\label{fig3}
\end{figure}

We also outperform previous fully automated and semi-automated methods on the task of RV endocardium and epicardium segmentation across all evaluation metrics. At test time, our model predicts endocardial and epicardial contours for both test1 and test2 sets (a total of 1,028 images) in less than a minute. Figure~\ref{fig3} illustrates some example endocardium and epicardium segmentation results for the right ventricle using the RVSC dataset. Again, note that fine-tuning the FCN model results in a significant accuracy boost over the same FCN model initialized with random features, thus establishing a new state of the art for right ventricle segmentation.

\begin{figure}[ht]
\centering
\includegraphics[width=\textwidth,height=\textheight,keepaspectratio]{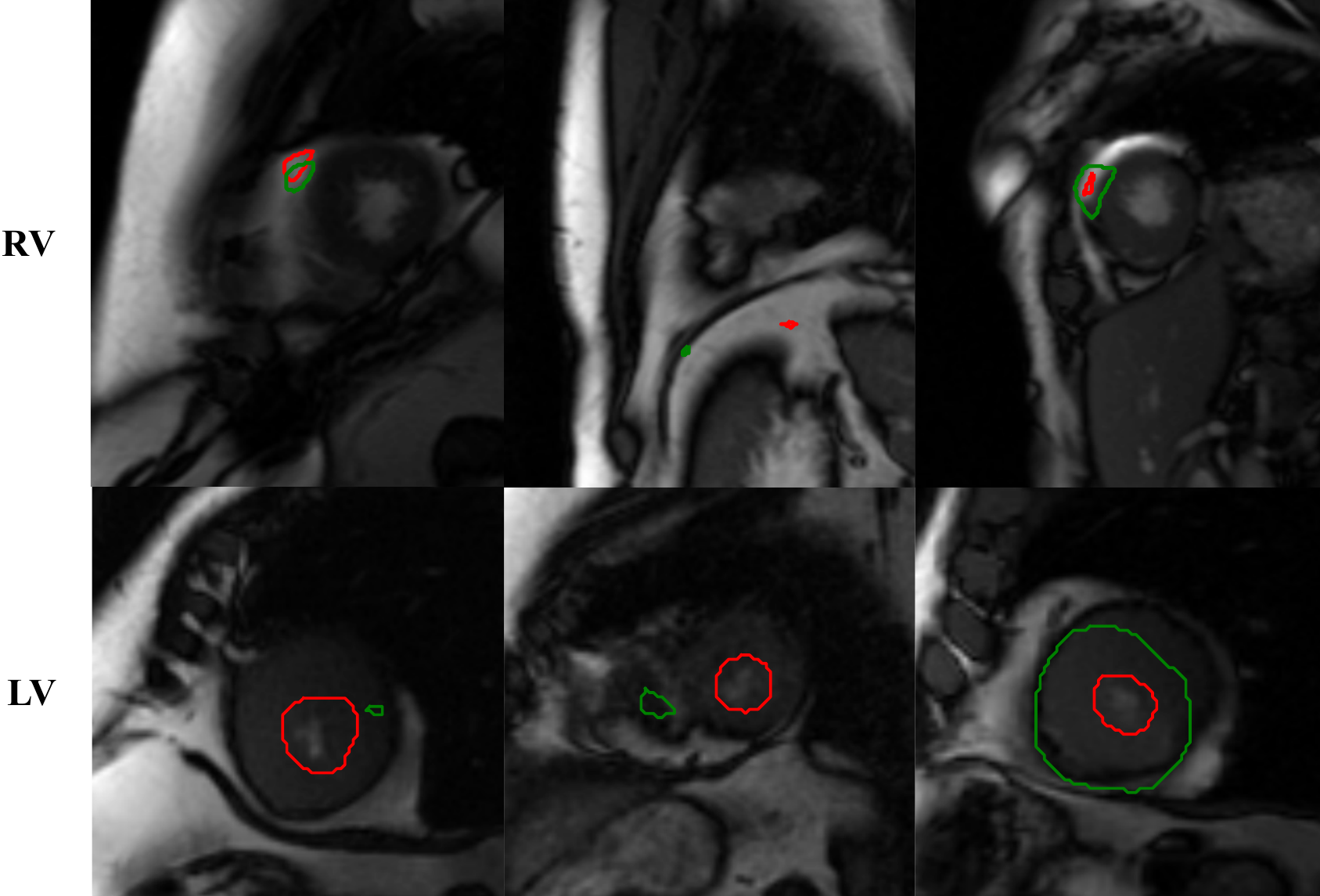}
\centering
\caption{Examples of poor FCN segmentation on difficult apical slices having ambiguous or imperceptible object boundaries. Cropped and zoomed in for better viewing. Colors: red -- endocardium; green -- epicardium.}
\label{fig4}
\end{figure}

Overall, the main limitation of the proposed FCN model lies in its inability to segment cardiac objects in difficult slices of the heart, especially at the apex. \citet{Petitjean:2015} report that the accuracy of previous RV segmentation methods also depends on slice level. Their analysis reveals that error is most prominent in apical slices. For example, the Dice index for the endocardial contour decreases by 0.20 from base to apex. **This analysis is also consistent in left ventricle segmentation**, where our FCN model fails to detect the presence of the cardiac object at the apex in several cases. Figure~\ref{fig4} shows some examples of poor FCN segmentation on these difficult apical slices. \citet{Petitjean:2015} suggest that the improvement of segmentation accuracy could be searched in apical slices, by emphasizing the model over the image content for these slices. Segmentation error on apical slices has minor impact on the volume computation, but it can be a limiting factor in other research fields such as the study of fiber structure \citep{Petitjean:2015}.


\section{Conclusion}
In this paper, we demonstrated the utility and efficacy of a fully convolutional neural network architecture for semantic segmentation in cardiac MRI. We showed that a single FCN model can be trained end-to-end to learn intricate features useful for segmenting both the left \emph{and} right ventricle. Comprehensive empirical evaluations revealed that our FCN model achieves state-of-the-art segmentation accuracy on multiple metrics and benchmark MRI datasets exhibiting real-world variability in image quality and cardiac anatomical and functional characteristics across sites, institutions, scanners, populations, and heart conditions. Moreover, the FCN model is fast, and can run on commodity compute resources such as the GPU to enable cardiac segmentation at massive scales.

The proposed FCN model can be further improved, in light of discovered limitations. The power of the FCN model lies in its capacity to learn millions of parameters on an abundance of training data. In order to improve segmentation accuracy of cardiac objects in difficult heart locations that exhibit ambiguous or imperceptible object boundaries such as apical and basal slices, the research community could dedicate effort in collecting more labeled or annotated examples at these locations. The demonstrated potential of the proposed FCN model is merely the tip of the iceberg. With more cardiac data to feed and train powerful, large-scale networks, FCN models could become the workhorse in advancing automated cardiac segmentation toward clinical applications with speed, accuracy, and reliability.

\section*{Acknowledgement}
The author thanks Alex Newton for some helpful code and gracious reviewers for constructive feedback on the paper.

\bibliographystyle{ormsv080} 
\bibliography{refs}

%

\end{document}